\documentclass{article}
\usepackage{ijcai25}

\usepackage{times}
\usepackage{soul}
\usepackage{url}
\usepackage[hidelinks]{hyperref}
\usepackage[utf8]{inputenc}
\usepackage[small]{caption}
\usepackage{graphicx}
\usepackage{amsmath}
\usepackage{amsthm}
\usepackage{booktabs}
\usepackage{algorithm}
\usepackage{algorithmic}
\usepackage[switch]{lineno}

\title{Fairness-aware Interactive Target Variable Definition}

\author{
Dalia Gala$^1$
\and
Milo Phillips-Brown$^{2}$\and
Naman Goel$^1$\and
Carina Prunkl$^{3}$\and
Laura Alvarez Jubete$^4$\and
medb corcoran$^4$\And
Ray Eitel-Porter$^4$\\
\affiliations
$^1$University of Oxford\\
$^2$University of Edinburgh\\
$^3$Utrecht University\\
$^4$Accenture\\
}

\begin{document}

\maketitle

\begin{abstract}
    Machine learning requires defining one's target variable for predictions or decisions, a process that can have profound implications for fairness, since biases are often encoded in target variable definition itself, before any data collection or training. The downstream impacts of target variable definition must be taken into account in order to responsibly develop, deploy, and use the algorithmic systems. We propose FairTargetSim (FTS), an interactive and simulation-based approach for this. We demonstrate FTS using the example of algorithmic hiring, grounded in real-world data and user-defined target variables. FTS is open-source; it can be used by algorithm developers, non-technical stakeholders, researchers, and educators in a number of ways. FTS is available at: \url{http://tinyurl.com/ftsinterface}. The video accompanying this paper is here: \url{http://tinyurl.com/ijcaifts}.
\end{abstract}


\section{Motivation}

\noindent Machine learning requires translating real-world problems into numerical representations. Sometimes, the translation is straightforward---e.g. in predicting whether someone defaults on a loan. Other times, things are not so simple. When developing an algorithm to predict which job applicants will be good employees, for example, one must make precise the notion of a ``good'' employee. This is an ambiguous, subjective notion about which reasonable minds may disagree. How one translates this notion numerically---how one \emph{defines the target variable}---can have profound implications for fairness \cite{passibarocas2019}. Defining ``good'' employee one way rather than another may result, e.g. in fewer applicants being hired from certain demographics. These issues arise in many domains. For a college admissions algorithm, one must determine who counts as a ``good'' student; for a search engine, one must determine what counts as a ``good'' search result; etc. How these notions are defined may likewise have weighty implications for fairness: which university applicants are admitted \cite{kizilceclee2022}; which items appear at the top of search results \cite{phillipsbrownms}; etc. Target variable definition, then, is not a merely technical matter. Defining ``good'' employee, student, or search result is a value-laden process: it calls for close attention and transparency \cite{fazelpourdanks2021}. 

But all too often, target variables are defined without transparency or attention to fairness. On one hand, technical developers may take target variable definition as a given, focusing instead on issues such as data quality, variance, accuracy of predictions, etc. On the other hand, stakeholders who are not a part of the technical process---like (hiring) managers in non-technical roles, or those working in upper management---either do not understand, or are simply unaware of, the implications of target variable definition in algorithmic settings. There is thus a pressing need for the fairness implications of target variable definition to be understood---and foregrounded---for stakeholders of all kinds.

To help meet this need, we developed an \emph{interactive target variable simulator}, FairTargetSim (FTS): \url{http://tinyurl.com/ftsinterface}. FTS introduces its users to target variable definition, and reveals and explains its impact on fairness. FTS uses a case study: hiring algorithms. FTS invites the user to imagine that they are building a hiring algorithm, which mirrors a widely-used style of hiring algorithm based on psychometric tests. The user defines two target variables, using real-world psychometric test data from \cite{Jaffe_Kaluszka_Ng_Schafer_2022}. With these two definitions, FTS builds two corresponding models and gives visualizations of how the models and training data differ in matters of fairness and overall performance. 

FTS's code is public and freely available. Therefore, its use is not limited to hiring algorithms or to the dataset we use in our case study: it can be extended to uses beyond education, and to different datasets and models.

\section{FairTargetSim's Audience}

FTS is a valuable tool for a wide range of audiences. The first target audience is technical developers who often want to develop algorithms responsibly but have less understanding of non-algorithmic factors such as target variable definition. With FTS, they can better understand the behavior of their abstract algorithms under different target variable definitions. This technical audience may also have less control over non-algorithmic factors, and can use FTS to better advocate---to decision-makers with non-technical backgrounds---for responsible algorithmic development. This leads us to the second target audience: non-technical stakeholders: e.g. those who use algorithms for making decisions or those who are impacted by the decisions. When these stakeholders better understand the fairness implications of target variable definition, the way is paved for more responsible and accountable use of algorithms in the real world. The third target audience is educators. There is a pressing need for more responsible AI education and training in universities (\cite{groszetal2018}, \cite{kopecetalfc}), government, and the private sector \cite{eitelporter2021}. The ethical implications of technical issues can be challenging to explain to learners. FTS gives educators an accessible, hands-on way to illustrate them.

We emphasize that FTS illustrates not ``only'' the fairness implications of decisions about target variable definition. It also illustrates, more generally, the ethical implications of decisions at the intersection of technical and non-technical aspects of algorithmic development. While it is well understood among theorists that such decisions are value-laden (\cite{friedmannissenbaum1996}, \cite{JohnsonForthcoming-JOHAAV}), they often do not wear their ethical dimensions on their sleeves. FTS allows audiences of all kinds to see---through a simulated algorithmic system---such decisions for what they are.

\section{Related Work}

A wealth of research has established the importance of understanding and addressing the fairness implications of target variable definition---in algorithmic systems generally (\cite{passijackson2018}, \cite{obermeyer2019dissecting}, \cite{martinetal2020}, \cite{levyetal2021}, \cite{barocasetal2023}) and hiring algorithms specifically (\cite{bazgucernea2019}, \cite{raghavanetal2020}, \cite{tilmes2022}). 

A number of systems have been developed for practitioners---and in some cases, non-technical stakeholders---to understand, identify, and address algorithmic bias. We list just some, and note that various of them, like FTS, have a visualization element: 
\cite{tramer2017}, \cite{bellamyetal2018}, \cite{ribeiroetal2018}, \cite{cabreraetal2019}, \cite{microsoft2019}, \cite{saleiroetal2019}, \cite{ahanetal2020}, \cite{wexler2020}, \cite{johnsonetal2023}, \cite{liuetal2023}. FTS is an important addition to these systems because it is, to our knowledge, the only one that addresses target variable definition. 

Compared to previous demonstrations at IJCAI on related subjects (e.g. ~\cite{sokol2018glass,juan2021hive,yu2019fair,miguel2021putting,henderson2021certifai,baumann2023bias}), our demonstration will focus on the problem of fairness implications of target variable definition.

\begin{figure}[t]
\centering
\includegraphics[width=0.95\columnwidth]{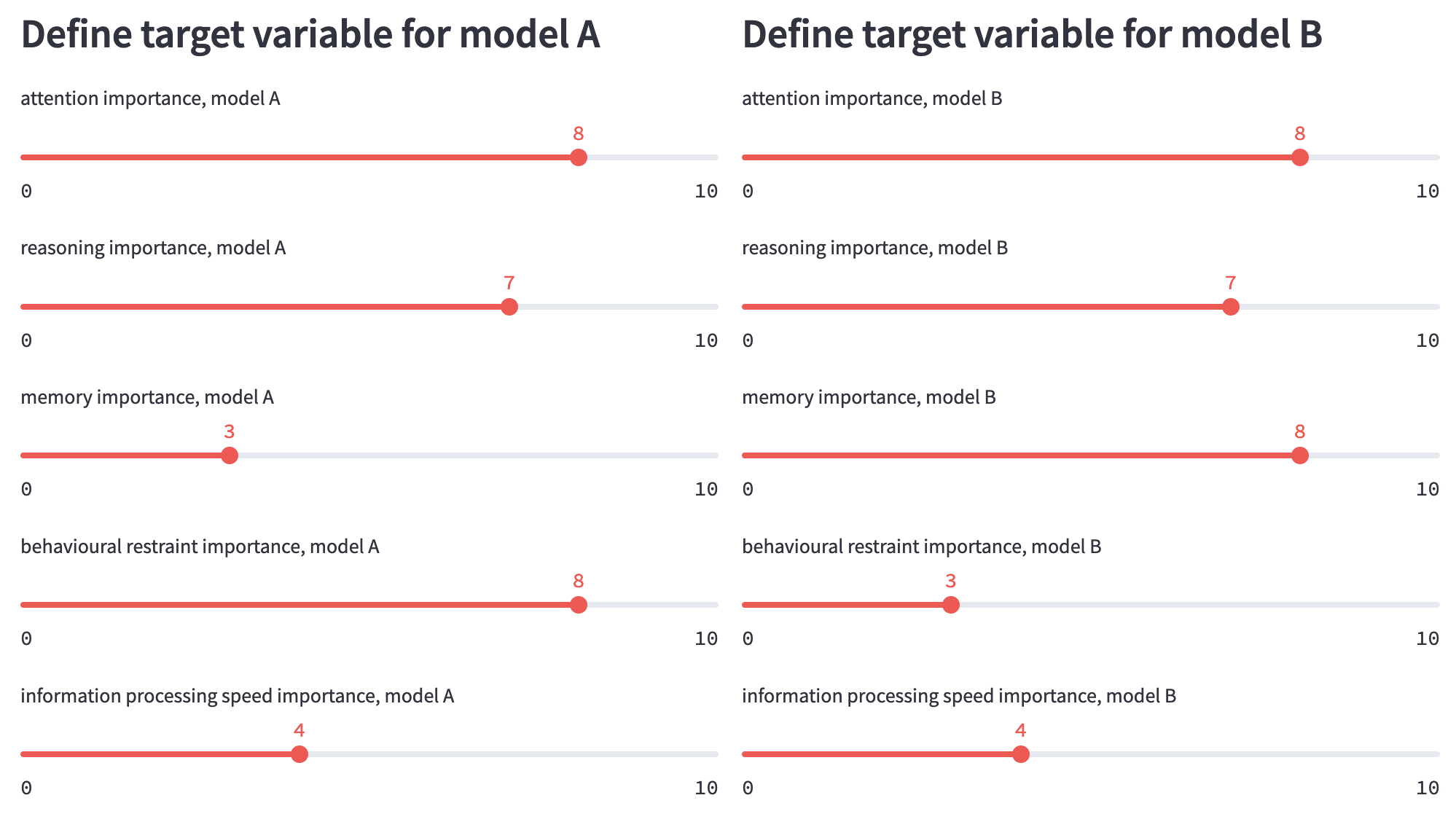} 
\caption{The user defines two target variables, using sliders representing the importance of traits of ``good'' employees.}
\label{fig1}
\end{figure}

\section{Overview of FairTargetSim}

FTS's interface works with most modern browsers; Firefox is advised. FTS has four pages that the user visits in order.

\subsection{Key Concepts Explained}

This page introduces target variable definition to a non-technical audience, explains how it impacts fairness, and gives an overview of the other pages of FTS.

\subsection{User Defines Target Variables}

This page has the user define two different target variables (Figure \ref{fig1}), which FTS uses to train two models, A and B. 

In the real-world hiring algorithms that are based in cognitive tests, developers often define ``good'' employee by having an employer identify a group of current employees whom the employer deems ``good'' for a given role \cite{wilsonetal2021}. These employees then play cognitive-test games, and a model is trained to identify applicants that share cognitive traits with these employees. 

FTS's models are similar to these real-world systems in two key ways. First, like those systems, FTS uses support vector machine models to identify people who share cognitive traits with those who are identified as ``good'' employees. Second, FTS's models are trained on data of real people's cognitive tests; the data we use is from Jaffe et al.'s (2022) battery 26, which has eleven tests that we grouped into five traits: memory, information-processing speed, reasoning, attention, and behavioral restraint.\footnote{Our five categories are based on the following tests: \emph{Memory} (forward memory span, reverse memory span, verbal list learning, delayed verbal list learning); \emph{Information Processing Speed} (digit symbol coding, trail making part A, trail-making part B); \emph{Reasoning} (arithmetic reasoning, grammatical reasoning); \emph{Attention} (divided visual attention); and \emph{Behavioral Restraint} (go/no-go).} 

FTS's models differ from the real-world systems in one key way: how the target variable is defined. With FTS, the user explicitly defines, using sliders depicted in Figure \ref{fig1}, how important the five cognitive traits are to what makes for a ``good employee.'' The user does this twice, creating two different target variables. Then FTS calculates the weighted average of test scores, given the slider weightings, and assigns class label ``0'' to those in the bottom 85th percentile. From the top 15\% subset, we randomly sample 100 ``good'' employees to whom we assign the class label ``1'' with weights ranging from 0.99 for the highest scoring candidate to 0.01 for the lowest scoring candidate, using linear distribution with the following equation for those in between: 

$$f(x)=\frac{0.98}{1-n}x +\frac{0.01-0.99n}{1-n}$$ 

We assign a class label ``0'' to those not selected, thus introducing randomness. FTS then generates two labeled datasets and corresponding models, each with different target variable definitions.

FTS works with user-defined target variables because, first, we do not have access to real-world target variables, and, second, the lessons FTS offers are brought to life for the user when she can see how her very own choices in target variable definition can have implications for fairness. As we explain further in Section \ref{sec:beyond}, having user-defined target variables is not a fundamental constraint on the idea of FTS; FTS can be extended to use real-world labels when they are available.

\subsection{Visualize Effects of Target Variable Definition}

This page contains visualizations that illustrate how the user's two target variable definitions impact issues of fairness and overall model performance. The visualizations are categorized into \emph{Demographics} and \emph{Non-demographics} sections, and further divided into categories that (i) show features of the models and (ii) features of the training data.

\begin{figure}[t]
\centering
\includegraphics[width=0.9\columnwidth]{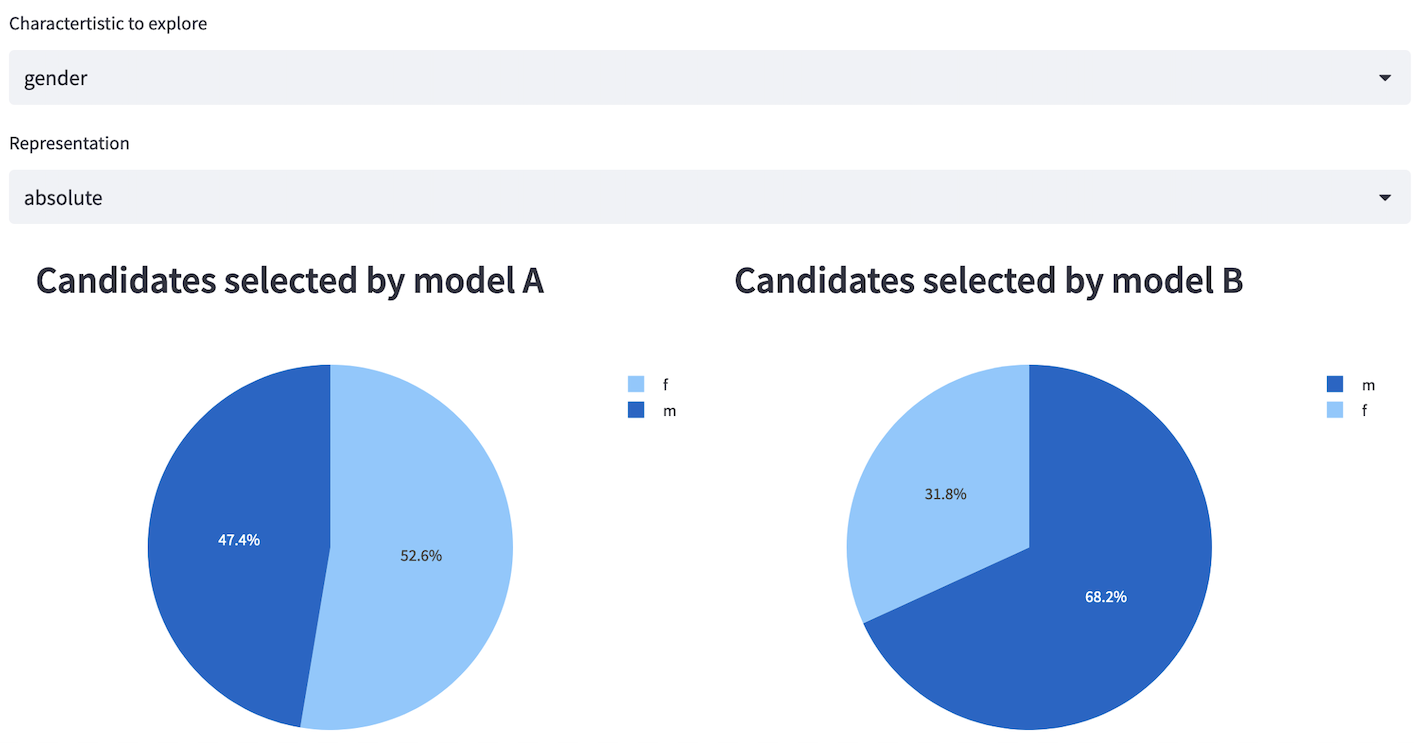}
\caption{Charts display how the percentage of selected male and female applicants differs between models A and B.}
\label{fig2}
\end{figure}

In the \emph{Demographic} section, charts as in Figure \ref{fig2} show how models A and B differ in, e.g. the proportions of selected applicants across demographic groups (gender, education level, age, and nationality---these are the demographic groups that the Jaffe \emph{et al.} dataset has information on). Other charts show how the models differ across groups with respect to ``fairness metrics'' (\cite{angwinetal2016}, \cite{corbettdaviesgoel2018}), such as true- and false-positive rates and positive and negative predictive value. 

The differences are stark: different target variable definitions often result in major differences in the demographics of selected applicants and in fairness metrics (see e.g. Figure 2). Visualizations in the \emph{Demographics} section also show how target variable definition affects models' training data: e.g. how positive and negative labels are distributed across demographic groups.

In the \emph{Non-demographic} section, visualizations show how the models and training data differ in ways other than fairness: e.g. how the models rank particular applicants (Figure 3), overall model confusion matrices, and accuracy metrics.

\begin{figure}[t]
\centering
\includegraphics[width=0.9\columnwidth]{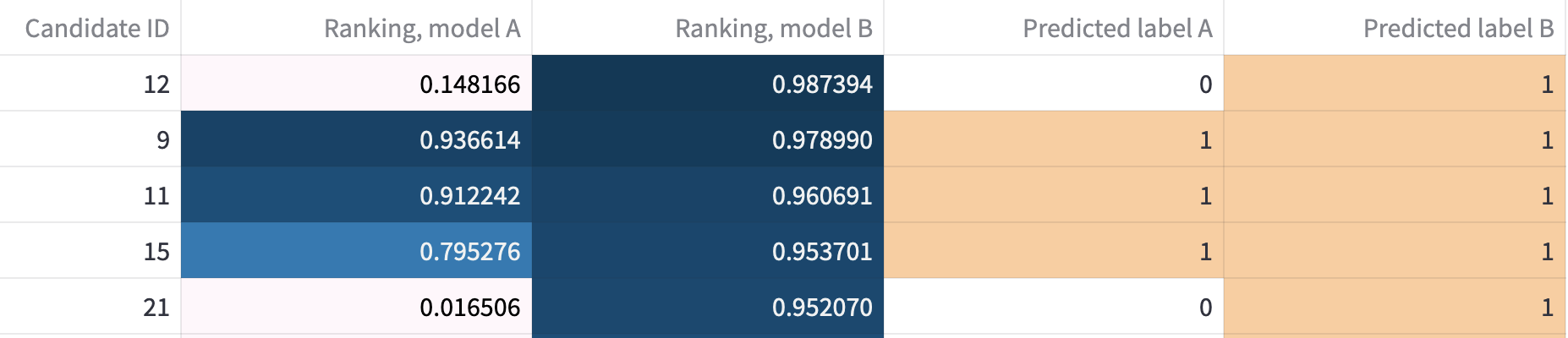} 
\caption{A table illustrates how individual applicants are evaluated differently by the two models.}
\label{fig3}
\end{figure}

\begin{figure}[h!]
\centering
\includegraphics[width=0.9\columnwidth]{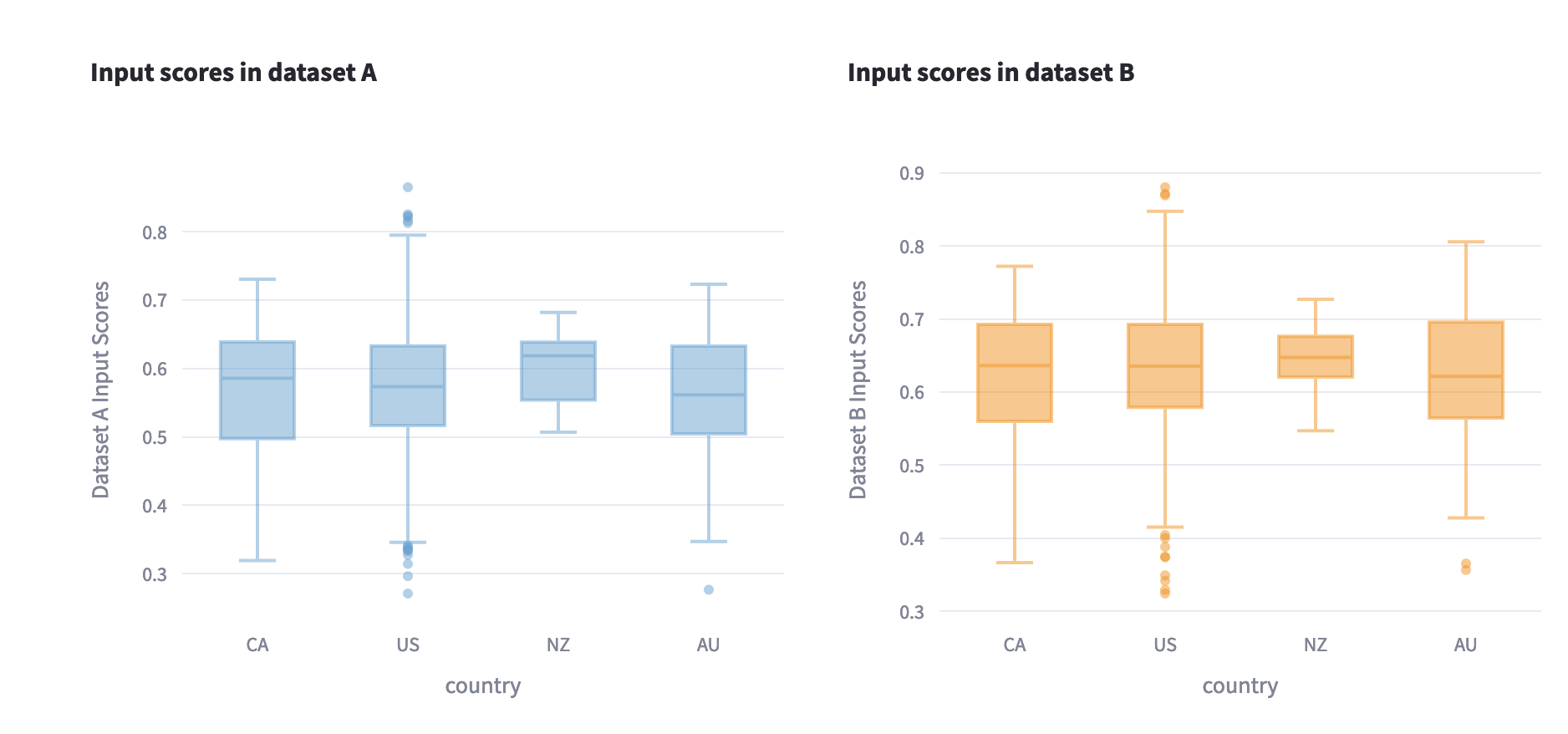} 
\caption{Bar graphs show how choice of features of importance affects the model input scores achieved for different candidates depending on the demographic group---in this case, country of origin. For example, for model A, the median score for American candidates is approximately 0.57, while for model B, it is 0.63.}
\label{fig3}
\end{figure}


\subsection{Further Uses of FairTargetSim}\label{sec:beyond}

This page gives recommendations for using FTS not just for providing explanations and educating stakeholders, but also for directly impacting practices in hiring and other domains. 

As noted, FTS's code is available publicly; an organization can extend FTS to use with their own data, models, and target variables. And, as also noted, in real-world target variable definition, employers do not directly identify cognitive characteristics of ``good'' employees; they identify certain current employees as ``good.'' We give guidance on how to do so in a way that can promote fairness. For example, (i) consult various managers on whom they judge ``good;'' these judgments can be weighted in different ways---just as FTS weights the cognitive tests in different ways---resulting in different target variables. Or, (ii) use various performance metrics to evaluate current employees (e.g. number of years to promotion, length of tenure at a company, or role-specific metrics, such as number of sales with a sales role); these metrics can, again, be weighted in different ways, resulting in different target variables. We also explain how to weight different judgements and metrics in other domains: 


\section{Future Work}

FTS opens up various avenues for future work, of which we will highlight a few. One, as noted in Section \ref{sec:beyond}, is to apply FTS to real-world hiring settings. Another, facilitated by the fact that FTS is flexible and openly available, is to invite the community to add more features to the simulator by, for example, using different kinds of datasets, models, or visualizations. Likewise, FTS could be extended to cases beyond algorithmic hiring, such as college admissions or search engines. 
Finally, FTS affords opportunities for human-centered research. For example, user-studies could be run---with both technical and non-technical stakeholders---to test how FTS affects how they think about, develop, and use algorithms for hiring and beyond.

\section*{Contribution Statement}

Gala and Phillips-Brown share first-authorship.


\bibliographystyle{named}
\bibliography{equivar}

\end{document}